\documentclass[conference]{IEEEtran}
\IEEEoverridecommandlockouts
\usepackage{tikz}
\usepackage{graphicx}
\usepackage{adjustbox}
\usepackage{booktabs}
\usepackage[noadjust]{cite}
\usepackage{amsmath,amssymb,amsfonts}
\usepackage{algorithmic}
\usepackage{graphicx}
\usepackage{textcomp}
\usepackage{xcolor}
\usepackage{array}
\usepackage{lipsum}

\renewcommand{\arraystretch}{2} 
\def\BibTeX{{\rm B\kern-.05em{\sc i\kern-.025em b}\kern-.08em
    T\kern-.1667em\lower.7ex\hbox{E}\kern-.125emX}}
\begin{document}

\title{Personalized Federated Learning for Cross-view Geo-localization

\thanks{This paper was supported by the the EU’s
H2020 TRUSTEE (No. 101070214).
}
}

\author{
\IEEEauthorblockN{Christos Anagnostopoulos$^{1,2}$, Alexandros Gkillas$^{1,2}$, Nikos Piperigkos$^{1,2}$,  Aris S. Lalos$^{1}$}
\IEEEauthorblockA{$^1$Industrial Systems Institute, Athena Research Center, Patras Science Park, Greece\\
$^2$AviSense.AI, Patras Science Park, Greece\\
Emails: \{anagnostopoulos, gkillas, piperigkos\}@avisense.ai, lalos@isi.gr
}
}

\newcommand\submittedtext{%
  \footnotesize \textcopyright 2024 IEEE. Personal use of this material is permitted. Permission from IEEE must be obtained for all other uses, in any current or future media, including reprinting/republishing this material for advertising or promotional purposes, creating new collective works, for resale or redistribution to servers or lists, or reuse of any copyrighted component of this work in other works. \linebreak Preprint submitted to the IEEE 26th International Workshop on Multimedia Signal Processing (MMSP).}

\newcommand\submittednotice{%
\begin{tikzpicture}[remember picture,overlay]
\node[anchor=south,yshift=10pt] at (current page.south) {\fbox{\parbox{\dimexpr0.65\textwidth-\fboxsep-\fboxrule\relax}{\submittedtext}}};
\end{tikzpicture}%
}
\maketitle
\submittednotice

\maketitle


\begin{abstract}
In this paper we propose a methodology combining Federated Learning (FL) with Cross-view Image Geo-localization (CVGL) techniques. We address the challenges of data privacy and heterogeneity in autonomous vehicle environments by proposing a personalized Federated Learning scenario that allows selective sharing of model parameters. Our method implements a coarse-to-fine approach, where clients share only the coarse feature extractors while keeping fine-grained features specific to local environments. We evaluate our approach against traditional centralized and single-client training schemes using the KITTI dataset combined with satellite imagery. Results demonstrate that our federated CVGL method achieves performance close to centralized training while maintaining data privacy. The proposed partial model sharing strategy shows comparable or slightly better performance than classical FL, offering significant reduced communication  overhead without sacrificing accuracy. Our work contributes to more robust and privacy-preserving localization systems for autonomous vehicles operating in diverse environments.
\end{abstract}

\begin{IEEEkeywords}
Federated Learning, Autonomous Driving, Cross-view, Geo-localization
\end{IEEEkeywords}

\section{Introduction}

Cross-view Geo-localization can provide a robust solution to the limitations of traditional GPS-based localization for vehicles. While GPS is generally reliable, it can fail in certain scenarios such as dense urban environments or when network connectivity is lost \cite{durgam2024crossview}. In these cases, relying solely on GPS can lead to inaccuracies and inconsistencies in vehicle localization. This can be accomplished by matching points of interest between the different views, the ground view and the satellite image. The arrival of Deep Learning (DL) techniques in feature extraction and their predominance over handcrafted techniques facilitated a new era characterized by improved accuracy and efficiency in processing and interpreting complex visual data, enabling more reliable Cross-view Geo-localization \cite{shi2022crossview}, 
\cite{fervers_uncertainty-aware_2022}.
\\
Data-driven models require large volumes of raw data for training something which can pose challenges advantages in terms of privacy and security \cite{9069945}. On the other hand, Federated Learning frameworks offer a promising solution by enabling models to be trained in a decentralized manner without the need to exchange local raw data, thus providing advantages in terms of data privacy and resource allocation \cite{zhang2021survey}.
\\
However, models participating in an FL training scheme often face challenges due to heterogeneous datasets, as autonomous vehicles typically operate in diverse environments \cite{lu2024federated}. To address this issue of local dataset heterogeneity, we propose a more personalized federated learning scenario. This approach implements a coarse-to-fine methodology, enforcing each client to selectively share only parts of its local model parameters. By doing so, we can better accommodate the unique characteristics of each vehicle's operating environment while still benefiting from collaborative learning. Furthermore, the aggregations of the components of local models responsible for extracting coarse, general features, which are more universal across different environments, can effectively improve overall model robustness and produce patterns that benefit all participants. Meanwhile, fine-grained feature extractors remain specific to each vehicle, capturing unique characteristics of local environments (e.g., urban vs. rural areas). This prevents potential performance degradation that could occur from aggregating highly specialized feature.
In addition, by selectively sharing only the coarse feature extractors, our approach not only enhances model performance but also improves computational efficiency by reducing the amount of data transmitted during the federated learning process. 
The main contributions of this paper can be summarized as follows:
\begin{itemize}
\item We propose the first methodology that combines federated learning with Cross-view image Geo-localization techniques. Our approach leverages deep neural networks to extract robust features from both satellite and ground-view images in a distributed setting, resulting in more accurate models that are resilient to view changes and discriminative for feature correspondences.

\item We introduce a personalized federated learning scenario to address dataset heterogeneity from autonomous vehicles in diverse environments. Our approach allows clients to selectively share parts of their local model parameters using a coarse-to-fine methodology. We aggregate components responsible for extracting general features, enhancing model robustness, while keeping fine-grained features specific to each vehicle to capture unique local characteristics. This prevents performance degradation from aggregating specialized features and improves computational efficiency by reducing data transmission.

%
    \item We conduct a thorough evaluation of the proposed federated learning Cross-view Geo-localization scenarios, comparing them against traditional centralized and single-client training schemes. This evaluation demonstrates the effectiveness of our approach.
    Extensive numerical results on a real-world dataset demonstrate the superiority of the proposed Federated Cross-view method and the benefits of cooperation to extract more robust features from diverse environments.
\end{itemize}
The remaining paper is structured as follows: Section II offers a brief overview of Cross-view Geo-localization and reviews related works on Federated Learning in the automotive domain. Section III details our methodology, including the theoretical formulation of the Cross View Geo Localization problem, the algorithm used, and the Federated Learning training schema. Section IV describes the experimental setup and presents the results. Finally, Section V concludes the paper and summarizes our findings.
\section{Preliminaries}
\subsection{Cross-view Geo-localization}

Cross-view Geo-localization addresses the problem of estimating a mobile agent's position, similar to Visual Odometry, but with a key distinction. While Visual Odometry matches image sequences from the same sensor, CVGL aims to match images from different sensors, and more specifically, the mobile agent's camera and geo-tagged aerial imagery. This approach leverages the complementary information provided by ground-level and aerial perspectives to determine the agent's location, offering a robust alternative to traditional localization methods, especially in scenarios where GPS might be unreliable or insufficient.

Research in this area typically focuses on two main problems, image retrieval techniques to match a ground view image to a geo-tagged aerial image, providing a rough estimate of the agent's position, and techniques using an initial rough pose estimation to determine the agent's pose with higher accuracy, which is also the problem we focus on in this study.
Initially, CVGL methods relied on pixel-wise techniques \cite{Sheikh2003}, \cite{6909903}. However, with the advent of feature-based methods, there was a shift towards related approaches \cite{Lin_2013_CVPR}. Following the rise of Deep Learning (DL) in computer vision, numerous DL-based works in CVGL have been presented, further advancing the field. Some implementations that can be found in the literature utilize technologies like Convolutional Neural Networks \cite{shi2022crossview}, \cite{shi2023boosting3dofgroundtosatellitecamera}, \cite{liu2019lendingorientationneuralnetworks},  Transformer methods \cite{fervers_uncertainty-aware_2022}, \cite{2204.00097}, \cite{https://doi.org/10.1049/cit2.12077}, and Generative Adversarial Networks based methods \cite{Coming}, \cite{2203.11832}.
\subsection{Federated Learning in Automotive Domain}

Although federated learning has been extensively explored in various fields such as image processing and computer vision, its application in the area of autonomous driving remains largely under-investigated. The current body of literature presents only a handful of studies investigating the potential benefits of federated learning in this domain. For instance, study \cite{9564384} utilized a federated learning framework to address object detection in automotive environments, achieving performance on par with centralized deep learning while enhancing the speed of local model training. Similarly, studies \cite{9533808} and \cite{9827020} utilized the FedAvg approach to develop precise local models for predicting wheel steering angles in autonomous vehicles.

Additionally, studies \cite{eldar} and \cite{9564384} delved into the theoretical aspects of federated learning in vehicular networks, examining the training processes of local models, data distribution challenges, and the non-i.i.d. characteristics of autonomous driving datasets. Studies in \cite{10314010, 10167942} provide a federated learning framework to tackle the lidar super-resolution problem, thus enhancing the quality of a low-cost  sensors utilizing lidar data from different autonomous vehicles under different environmental conditions.
Furthermore, in \cite{10167897}, the authors applied federated learning principles to train a deep learning-based feature detector, integrated it into a SLAM system and evaluated its performance using odometry metrics. Lastly, study \cite{9981098} focused on semantic segmentation through federated learning, establishing a benchmark platform that incorporated two datasets and multiple leading federated learning algorithms.
To our knowledge, this study is the first to combine Cross-view Geo-localization with Federated Learning, introducing a personalized FL scenario. In this approach, clients selectively share parts of the Deep Learning-based feature extractor using a coarse-to-fine methodology, aggregating only the components responsible for extracting general features.

\section{Methodology}
\subsection{CVGL Problem Formulation}
The problem of Cross-view Geo-localization can be classified under two types. The first type involves a ground view query image \( Q \) with unknown geographic coordinates, which must be matched against a set of aerial images \(\{I_1, I_2, \ldots, I_n\}\) that are geotagged, resulting in a coarse ego vehicle's position estimation. 
The second type, which is going to be examined in this study, uses an initial rough pose estimation and relates the ground view image with an already selected aerial image centered on the initial pose, resulting in predictions far more accurate in terms of translation and rotation. 
\\
The relation of a pixel in the ground view image, given that it uses a pinhole camera model, and the a point in the camera's world coordinate system is 

\[
\tilde{\mathbf{x}}_g = P \tilde{\mathbf{X}}_w
\]

where $\tilde{\mathbf{x}}_g$ and $\tilde{\mathbf{X}}_w$ are the positions of the point in ground view image and the world using homogeneous coordinates,  \( P = K (R^\top - R^\top t) \) is the camera matrix, \( K \) is the camera's calibration matrix, and \( R \), \( t \) are the camera's rotation and translation in the world coordinate system.
In addition $\tilde{\mathbf{X}}_w$ can be mapped to a pixel in the aerial image \(\tilde{\mathbf{x}_s} = f(\tilde{\mathbf{X}_w}, a)\), where \( f: \mathbb{R}^3 \to \mathbb{R}^2 \) is the projection function to the aerial image plane, $\tilde{\mathbf{x}_s}$ is the position of the point in the aerial image, and $a$ is the real world distance of each pixel of the aerial map. 
Therefore, we define a projective relationship between a point in the ground view and a point in the satellite image, which includes the position and rotation of the camera in world coordinates. 
\\
Our goal is to find the optimal pose $\hat{R}$ and $\hat{t}$ of the ground camera by minimizing a loss function:
\begin{equation}
\hat{\xi} = \arg \min_{\xi} \|e\|^*
\label{eq:loss_function}
\end{equation}
where $\|e\|^*$ is a generic distance measure between the aerial and the projected observations from the ground view image, $\xi = \{R, t\}$, and $\hat{\xi}$ corresponds to its optimal solution.

\subsection{Federated Learning CVGL}

To define the CVGL federated learning problem, we consider a network with 
$N$ autonomous vehicles (agents), each possessing a local private dataset of size $k_j$ which can be defined as:
\[
D_j = \{I^s_i, I^g_i, R^*_i, t^*_i\}
\]
where \( j \in \{1, \ldots, N\} \), \( i \in \{1, \ldots, k_j\} \), \( I^s_i \in \mathbb{R}^{ H_{s} \times W_{s} \times C_{s}} \) and  \( I^g_i \in \mathbb{R}^{ H_{g} \times W_{g} \times C_{g}} \) represent the satellite images and the ground view images, \(R^*_i \in \mathbb{R}^{3 \times 3}\) be the ground truth rotation matrix and  and \(t^*_i \in \mathbb{R}^3\) represents the ground truth translations.
Here, \(H_{s}\) and \(W_{s}\) are the height and width of the satellite images, respectively, and \(C_{s}\) is the number of channels. Similarly, \(H_{g}\) and \(W_{g}\) are the height and width of the ground view images, respectively, and \(C_{g}\) is the number of channels.
\\
 Within the federated learning framework, these devices collaboratively train a global model, orchestrated by a central server, which minimizes the aggregation of local objectives. 

\subsubsection{Client side}
\begin{figure}
\centering
 \includegraphics[scale=0.31]{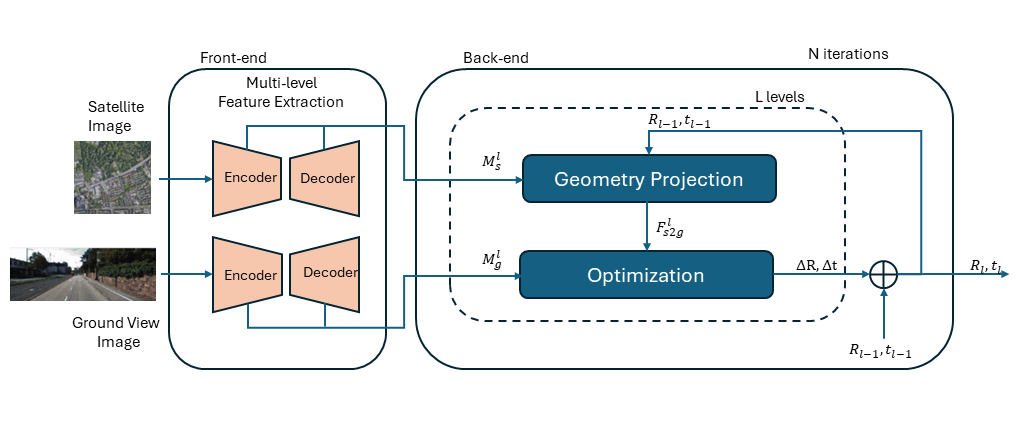}
 \caption{The client's model is divided into two main components: the front-end and the back-end. The front-end features two parallel VGG-based branches following an autoencoder architecture to extract multi-scale feature maps from satellite and ground view images. The extracted features are then projected to the ground view plane to establish cross-view geometric correspondences. The back-end performs an optimization process using the Levenberg-Marquardt (LM) algorithm, iteratively, for each  refining the pose estimation from the coarsest to finer feature levels until convergence or a maximum of five iterations is reached.}
    \label{fig:system}
\end{figure}

Each client $n$ utilizes its privately owned dataset $D_j$ to optimize a local deep learning-based CVGL model based on the work presented in \cite{shi_beyond_2022}. 
More specifically, the model illustrated in Figure \ref{fig:system} can be divided into two components, similar to a typical vSLAM system: the front-end and the back-end. 
\\
The front-end consists of two parallel VGG-based branches \cite{vgg} following an autoencoder architecture. Although these deep learning-based extractors are identical in structure, they do not share the same weights, allowing them to better adapt to their respective domains. In order to be less sensitive to object distortions we employ feature maps, which can encode high-level information, instead of RGB values.
Therefore, both branches are responsible for simultaneously extracting multiple feature maps from different layers of the networks, denoted as $\mathbf{M}_{s}^l \in \mathbb{R}^{H_{s}^l \times W_{s}^l \times C^l_s}$ for the satellite images and $\mathbf{M}_{g}^l \in \mathbb{R}^{H_{g}^l \times W_{g}^l \times C^l_g}$ for the ground view images, where $l$ indicates the layer of the network. 
\\
This integration of features from different layers ensures that the model benefits from a rich, multi-scale feature representation, which is essential for effectively processing and correlating satellite and ground view images. Subsequently, the extracted satellite features are projected to the ground view plane, resulting to a new feature map $M_{s2g}^l \in \mathbb{R}^{H_{g}^l \times W_{g}^l \times C^l}$, and subsequently the local loss function for each client $j$, becomes
\begin{equation}
\mathbf{e}^l_j = \mathbf{M}_{s2g}^l - \mathbf{M}_{g}^l
\end{equation}
The projection function depends on the pose of the camera  and uses the homography of the ground plane making the approximation that the majority of the points between the two views lie on the ground plane. 
\\

\begin{figure*}
\centering
 \includegraphics[scale=0.35]{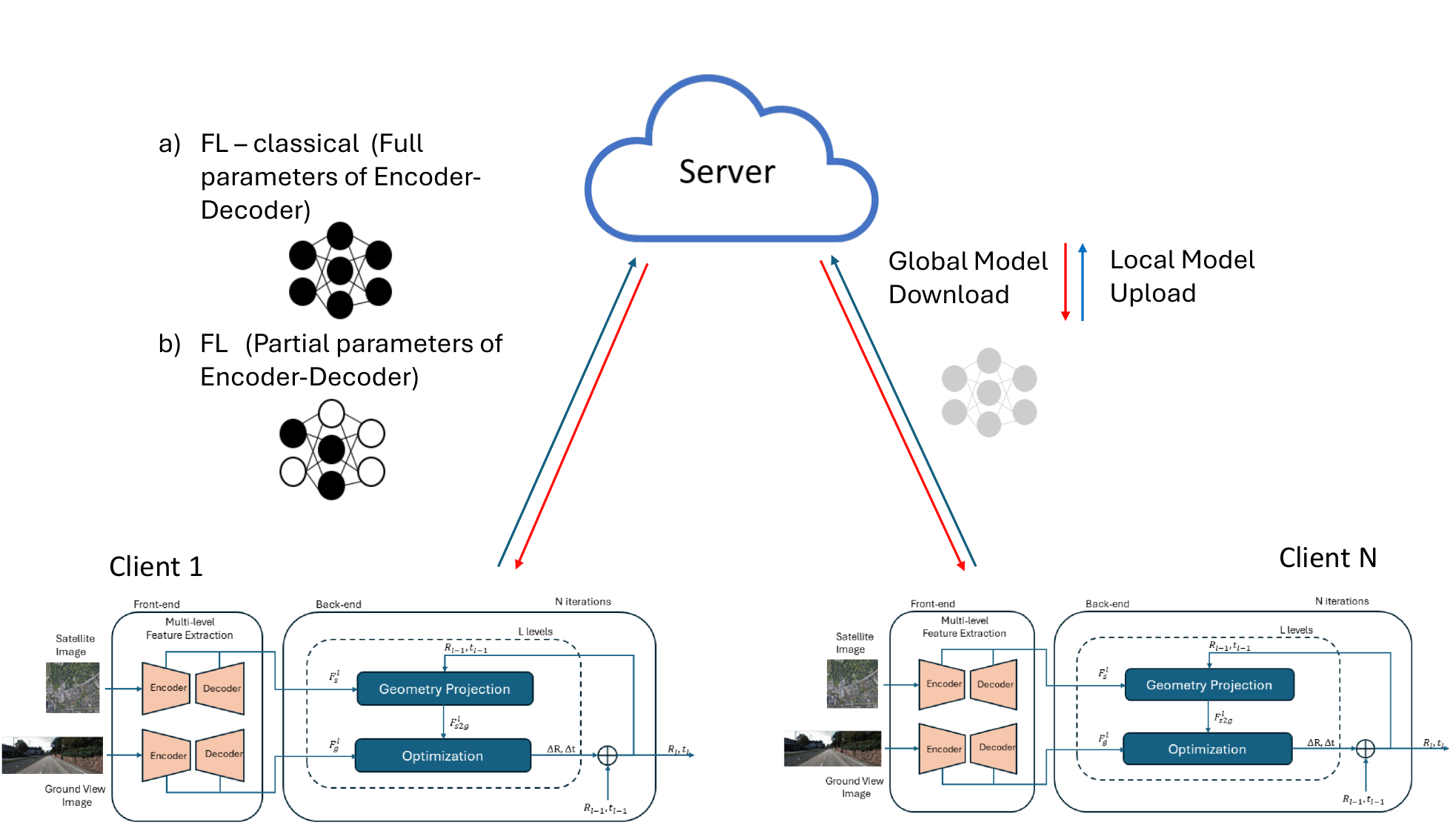}
 \caption{Overview of the Federated Learning Framework that consists of N clients, each using their private datasets to refine a local deep learning-based CVGL model. We consider two scenarios for model exchange:
(a) Full Model Sharing, where he entire model is exchanged between clients and (b)  Partial Encoder Sharing where the clients share only the encoders of their local model parameters, employing a coarse-to-fine approach.}
    \label{fig:FLsystem}
\end{figure*}

The backend runs an optimization process that refines the estimated pose, ensuring precise alignment between the ground and satellite images through non-linear optimization that minimizes the difference between projected points. This implementation employs the Levenberg-Marquardt (LM) \cite{LM} optimization algorithm to solve the resulting non-linear least squares problem. 
The LM optimization is implemented in a differentiable
manner, operating in a feed-forward pass in a coarse-to fine methodology. At each level \(l\), it is computed the Jacobian and the Hessian matrix
\begin{equation}
    \mathbf{J}^l_j = \frac{\partial \mathbf{M}^l_{s2g}}{\partial \xi} = \frac{\partial \mathbf{M}^l_{s2g}}{\partial \mathbf{p}_s} \frac{\partial \mathbf{p}_s}{\partial \xi_j^l}, \mathbf{H}^l_j =(\mathbf{J}^l_j)^{\top} \mathbf{J}^l_j
\end{equation}where \(\mathbf{p}_s\) are satellite feature map coordinates. 
In each algorithmic iteration we have as small pose update $\Delta \xi^l_j = (\mathbf{H}_j^l)^{-1} (\mathbf{J}^l_j)^{\top}\mathbf{e}^l_j$ which results in an updated pose:
\begin{equation}
    \xi^{l+1}_j = \xi^l_j + \Delta \xi^l_j
\end{equation}
where $\Delta \xi^l_j =  \{\Delta R_j^l, \Delta t_j^l\}$. The optimization is applied at the coarsest feature level and then progressively advances to finer levels, resulting in an optimization process executed as many times as the number of feature maps extracted for each view. The inputs of each step are the corresponding feature maps for the layer $l+1$ and the estimated rotation and translation from the previous layer $\xi_j^l$. This is repeated iteratively until convergence is achieved or a maximum of $\tau$ iterations is reached.
\\
The network is trained end-to-end using ground truth (GT) camera poses for supervision. The loss function is defined as follows:
\begin{equation}
\mathcal{L}_j = \sum_{t} \sum_{l} \left( \| \hat{\mathbf{R}}_{j,t}^l - \mathbf{R}^*_j \| + \| \hat{\mathbf{t}}_{j,t}^l - \mathbf{t}^*_j \| \right),
\label{eq:generic_loss}
\end{equation}
where $\hat{\mathbf{R}}_{j,t}^l$ and $\hat{\mathbf{j,t}}_t^l$ are the predicted poses  at the $t$-th iteration and $l$-th level, respectively, and $\mathbf{R}^*_j$ and $\mathbf{t}^*_j$ are the GT camera poses for client $j$.

\subsubsection{Server-side}
On the server-side, the primary objective is to compute a global model by integrating the local models received from the participating autonomous agents as illustrated in Figure \ref{fig:FLsystem}. Specifically, the server aggregates the local models to produce a new global model, denoted as  $\vartheta_g$, through a weighted average fusion method \cite{mcmahan2023communicationefficientlearningdeepnetworks}.
The mathematical formulation for this process is as follows:
\begin{equation}
\vartheta_g = \frac{1}{N}\sum_{j=1}^N w_j \vartheta_j,
\label{eq}
\end{equation}

In equation \eqref{eq}, $w_j$ represents the weight corresponding to the size of the local dataset of the $j$-th device. This weight ensures that models trained on larger datasets have a proportionately greater influence on the global model. The weighted average fusion approach is critical as it balances the contribution of each local model, taking into account the variability in dataset sizes across different devices.
After the local models are merged using this method, the centralized server disseminates the updated global model back to all participating devices. This distribution allows each device to update its local model with the newly computed global parameters, ensuring that all agents benefit from the collective training process.
\\
The process of aggregating and distributing the global model is repeated for T communication rounds. During each round, local models are trained, aggregated, and redistributed, progressively refining the global model. This iterative process continues until the global model converges to an optimal solution.
\\
We consider two scenarios for model exchange: (a) Full Model
Sharing, where he entire model is exchanged between clients and (b) Partial Encoder Sharing where the clients share only the encoders of their local model parameters, employing a coarse-to-fine approach aiming at the adaption to diverse data distributions, and the optimization of computational resources.

\begin{table*}
\renewcommand{\arraystretch}{2} 
\caption{Maximum values for each scenario and each individual metric.}
\label{tab:scenario_metrics_non_combined}
\centering
\begin{adjustbox}{max width=\textwidth}
\begin{tabular}{|l||c|c|c|c|c|c|c|c|c|}
\hline
\textbf{Scenario} & \textbf{Distance 1m} & \textbf{Distance 3m} & \textbf{Distance 5m} & \textbf{Lateral 1m} & \textbf{Lateral 3m} & \textbf{Lateral 5m} & \textbf{Angle 1°} & \textbf{Angle 3°} & \textbf{Angle 5°} \\
\hline
Single Client & 1.18 & 7.61 & 16.3 & 22.89 & 54.11 & 67.79 & 16.14 & 42.77 & 63.3 \\
\hline
FL (classical) & 1.56 & 10.13 & 20.07 & 31.42 & 67.14 & 76.86 & 21.81 & 54.64 & 74.48 \\
\hline

FL (Only Encoders) & 1.55 & 9.98 & 19.74 & 31.23 & 67 & 76.86 & 22.08 & 57.69 & 76.09 \\
\hline
Centralized & \textbf{2.09} & \textbf{11.59} & \textbf{20.58} & \textbf{34.86} & \textbf{67.79} & \textbf{77.62} & \textbf{23.6} & \textbf{59.49} & \textbf{77.42} \\
\hline
\end{tabular}
\end{adjustbox}
\end{table*}

\begin{table*}
\renewcommand{\arraystretch}{2} 
\caption{Maximum values for combined metrics for each scenario.}
\label{tab:scenario_metrics_combined}
\centering
\begin{tabular}{|l||c|c|c|}
\hline
\textbf{Scenario} & \textbf{Lat 1m \& Angle 1°} & \textbf{Lat 3m \& Angle 3°} & \textbf{Lat 5m \& Angle 5°} \\
\hline
Single Client & 5.01 & 26.66 & 45.8 \\
\hline
FL (Classical) & 9.93 & 42.46 & 61.24 \\
\hline

FL (Only Encoders) & 9.94 & 43.86 & 62.76 \\
\hline

Centralized & \textbf{11.34} & \textbf{46.83} & \textbf{63.95} \\
\hline
\end{tabular}

\end{table*}

\section{Experiments}
\subsection{Dataset}
We used the KITTI raw dataset \cite{Geiger2013IJRR} combined with satellite images retrieved from Google Maps, following the procedure described in \cite{shi2022crossview}. The dataset was split into five parts, each containing over 3000 images with non-overlapping sequences. The testing dataset comprises images from a different area to ensure generalization, allowing us to evaluate the model's performance in unseen environments.
\subsection{Metrics}
We use the metrics proposed in \cite{shi2022crossview} for the evaluation of the estimated pose in the lateral and longitudinal directions as well as orientation. An estimation is considered successful if the value is below certain thresholds. Specifically, for translation, in the lateral or longitudinal direction, the threshold \(d\) is set to 1, 3, and 5 meters, and for the azimuth angle, the threshold is set to 1°, 3°, and 5°. The evaluation also includes combined metrics, such as latitude below 1 meter and azimuth below 1 degree.

\subsection{Experimental Setup}
In the first two scenarios, we examined two extreme cases where both models followed traditional centralized training but with different data accessibility. The first model had access to all data from all agents, while the second model had access to data from only one agent. The other two scenarios incorporated FL principles and were expected to perform between these two extreme cases, at least in terms of accuracy.
The scenarios can be summarized in the following list:
\begin{itemize}
    \item \textbf{Centralized Training}: The entire dataset was used for training, something that is expected to produce the best results. 
    \item \textbf{Single Client Training}: Each client had access only to their respective dataset partition for training thus limiting their ability to produce a model that could generalize to a different dataset and hindering their feature extracting capability.
    \item \textbf{Federated Learning}: Classical Federated Learning Training scheme where each client sends the whole trained model.
    \item \textbf{Federated Learning with Partial Encoder Exchange}: To address the challenges of data heterogeneity and communication overhead in FL for CVGL, we explore the partial exchange of the VGG-based feature extractors for both satellite and ground views.
\end{itemize}
In both FL cases the generated global models were trained for one epoch for finetuning and for seven communication rounds. All the experiments were conducted on a platform equipped with an NVIDIA RTX 2080 Ti GPU, an Intel Core i7-8700 CPU, and 32GB of RAM.

\subsection{Evaluation Study}
The experimental results demonstrate clear performance differences across the four scenarios tested. As expected, the Centralized Training scenario  outperforms the other approaches across all metrics. This is evident in both individual metrics, as shown in Table \ref{tab:scenario_metrics_combined}, and combined metrics that are depicted in Table \ref{tab:scenario_metrics_non_combined}. Nonetheless, centralized training necessitates access to the entire dataset, leading to potential privacy concerns and increased communication burdens due to the requirement for clients to upload large volumes of raw data.
\begin{figure}
\centering
 \includegraphics[scale=0.44]{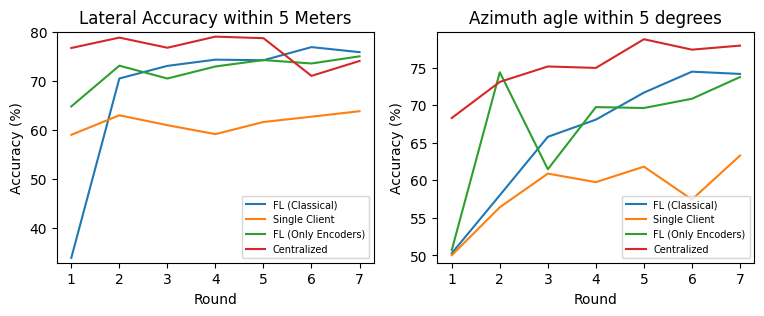}
 \caption{Comparison of accuracy across different scenarios and two different metrics. The left image illustrates the lateral accuracy within 5 meters and the right image the azimuth accuracy within five degrees.}
    \label{fig:combined_accuracy}
\end{figure}

Interestingly, both Federated Learning approaches (Classical FL and FL with Only Encoders) show comparable performance, often achieving results close to the Centralized Training scenario. This is particularly evident, as shown in Figure \ref{fig:combined_accuracy}  in the lateral accuracy within 5 meters and the azimuth angle within 5 degrees, where in especially in the first case both FL approaches nearly match the Centralized Training performance by the final round. The FL (Only Encoders) approach even slightly outperforms the Classical FL in some metrics, such as the azimuth angle accuracy (76.09\% vs 74.48\% within 5°), suggesting that sharing only the encoder parameters can be as effective as sharing the entire model in this context. Thus, you achieve similar accuracy while exchanging only approximately 50\% of the model parameters, potentially reducing the communication overhead significantly.

The Single Client scenario consistently shows the lowest performance across all metrics, highlighting the limitations of training with a restricted dataset. However, it's worth noting that the performance gap between the Single Client and the FL approaches is substantial, demonstrating the clear benefits of federated learning in improving model accuracy without centralizing data. This is particularly evident in the azimuth angle accuracy within 5 degrees, where both FL approaches show significant improvement over the Single Client scenario, especially in later rounds. These results underscore the potential of federated learning to balance data privacy concerns with the need for robust models in Cross-view Geo-localization tasks.



\section{Conclusion}
This paper presents a novel combination of Federated Learning  with Cross-view Geo-localization to address the limitations of GPS-based localization in autonomous vehicles. The proposed coarse-to-fine approach show significant improvements over single-client training and perform close to centralized training. This highlights the potential of FL to balance data privacy concerns with the need for robust, generalizable models.  Overall, our work contributes to the advancement of collaborative learning in autonomous driving systems, paving the way for more robust and adaptable localization methods. Future work could explore  the scalability of the approach with larger numbers of clients, and the integration of this method with other sensors and localization techniques.

\bibliographystyle{IEEEtran}
\bibliography{MMSP2024_FL.bib}

\begin{thebibliography}{10}
\providecommand{\url}[1]{#1}
\csname url@samestyle\endcsname
\providecommand{\newblock}{\relax}
\providecommand{\bibinfo}[2]{#2}
\providecommand{\BIBentrySTDinterwordspacing}{\spaceskip=0pt\relax}
\providecommand{\BIBentryALTinterwordstretchfactor}{4}
\providecommand{\BIBentryALTinterwordspacing}{\spaceskip=\fontdimen2\font plus
\BIBentryALTinterwordstretchfactor\fontdimen3\font minus \fontdimen4\font\relax}
\providecommand{\BIBforeignlanguage}[2]{{%
\expandafter\ifx\csname l@#1\endcsname\relax
\typeout{** WARNING: IEEEtran.bst: No hyphenation pattern has been}%
\typeout{** loaded for the language `#1'. Using the pattern for}%
\typeout{** the default language instead.}%
\else
\language=\csname l@#1\endcsname
\fi
#2}}
\providecommand{\BIBdecl}{\relax}
\BIBdecl

\bibitem{durgam2024crossview}
A.~Durgam, S.~Paheding, V.~Dhiman, and V.~Devabhaktuni, ``Cross-view geo-localization: a survey,'' 2024.

\bibitem{shi2022crossview}
Y.~Shi and H.~Li, ``Beyond cross-view image retrieval: Highly accurate vehicle localization using satellite image,'' 2022.

\bibitem{fervers_uncertainty-aware_2022}
\BIBentryALTinterwordspacing
F.~Fervers, S.~Bullinger, C.~Bodensteiner, M.~Arens, and R.~Stiefelhagen, ``Uncertainty-aware {Vision}-based {Metric} {Cross}-view {Geolocalization},'' 2022, version Number: 2. [Online]. Available: \url{https://arxiv.org/abs/2211.12145}
\BIBentrySTDinterwordspacing

\bibitem{9069945}
K.~Wei, J.~Li, M.~Ding, C.~Ma, H.~H. Yang, F.~Farokhi, S.~Jin, T.~Q.~S. Quek, and H.~Vincent~Poor, ``Federated learning with differential privacy: Algorithms and performance analysis,'' \emph{IEEE Transactions on Information Forensics and Security}, vol.~15, pp. 3454--3469, 2020.

\bibitem{zhang2021survey}
C.~Zhang, Y.~Xie, H.~Bai, B.~Yu, W.~Li, and Y.~Gao, ``A survey on federated learning,'' \emph{Knowledge-Based Systems}, vol. 216, p. 106775, 2021.

\bibitem{lu2024federated}
Z.~Lu, H.~Pan, Y.~Dai, X.~Si, and Y.~Zhang, ``Federated learning with non-iid data: A survey,'' \emph{IEEE Internet of Things Journal}, 2024.

\bibitem{Sheikh2003}
Y.~Sheikh, S.~Khan, M.~Shah, and R.~Cannata, \emph{Geodetic Alignment of Aerial Video Frames}.\hskip 1em plus 0.5em minus 0.4em\relax Boston, MA: Springer US, 2003, pp. 144--179.

\bibitem{6909903}
M.~Bansal and K.~Daniilidis, ``Geometric urban geo-localization,'' in \emph{2014 IEEE Conference on Computer Vision and Pattern Recognition}, 2014, pp. 3978--3985.

\bibitem{Lin_2013_CVPR}
T.-Y. Lin, S.~Belongie, and J.~Hays, ``Cross-view image geolocalization,'' in \emph{Proceedings of the IEEE Conference on Computer Vision and Pattern Recognition (CVPR)}, June 2013.

\bibitem{shi2023boosting3dofgroundtosatellitecamera}
\BIBentryALTinterwordspacing
Y.~Shi, F.~Wu, A.~Perincherry, A.~Vora, and H.~Li, ``Boosting 3-dof ground-to-satellite camera localization accuracy via geometry-guided cross-view transformer,'' 2023. [Online]. Available: \url{https://arxiv.org/abs/2307.08015}
\BIBentrySTDinterwordspacing

\bibitem{liu2019lendingorientationneuralnetworks}
L.~Liu and H.~Li, ``Lending orientation to neural networks for cross-view geo-localization,'' 2019.

\bibitem{2204.00097}
S.~Zhu, M.~Shah, and C.~Chen, ``Transgeo: Transformer is all you need for cross-view image geo-localization,'' 2022.

\bibitem{https://doi.org/10.1049/cit2.12077}
\BIBentryALTinterwordspacing
S.~Li, Z.~Tu, Y.~Chen, and T.~Yu, ``Multi-scale attention encoder for street-to-aerial image geo-localization,'' \emph{CAAI Transactions on Intelligence Technology}, vol.~8, no.~1, pp. 166--176, 2023. [Online]. Available: \url{https://ietresearch.onlinelibrary.wiley.com/doi/abs/10.1049/cit2.12077}
\BIBentrySTDinterwordspacing

\bibitem{Coming}
\BIBentryALTinterwordspacing
A.~Toker, Q.~Zhou, M.~Maximov, and L.~Leal-Taixé, ``Coming down to earth: Satellite-to-street view synthesis for geo-localization,'' 2021. [Online]. Available: \url{https://arxiv.org/abs/2103.06818}
\BIBentrySTDinterwordspacing

\bibitem{2203.11832}
S.~Wu, H.~Tang, X.-Y. Jing, H.~Zhao, J.~Qian, N.~Sebe, and Y.~Yan, ``Cross-view panorama image synthesis,'' 2022.

\bibitem{9564384}
D.~Jallepalli, N.~C. Ravikumar, P.~V. Badarinath, S.~Uchil, and M.~A. Suresh, ``Federated learning for object detection in autonomous vehicles,'' in \emph{2021 IEEE Seventh International Conference on Big Data Computing Service and Applications (BigDataService)}, 2021, pp. 107--114.

\bibitem{9533808}
H.~Zhang, J.~Bosch, and H.~H. Olsson, ``End-to-end federated learning for autonomous driving vehicles,'' in \emph{2021 International Joint Conference on Neural Networks (IJCNN)}, 2021, pp. 1--8.

\bibitem{9827020}
A.~Nguyen, T.~Do, M.~Tran, B.~X. Nguyen, C.~Duong, T.~Phan, E.~Tjiputra, and Q.~D. Tran, ``Deep federated learning for autonomous driving,'' in \emph{2022 IEEE Intelligent Vehicles Symposium (IV)}, 2022, pp. 1824--1830.

\bibitem{eldar}
S.~Wang, C.~Li, D.~W.~K. Ng, Y.~C. Eldar, H.~V. Poor, Q.~Hao, and C.~Xu, ``Federated deep learning meets autonomous vehicle perception: Design and verification,'' \emph{IEEE Network}, pp. 1--10, 2022.

\bibitem{10314010}
A.~Gkillas, A.~S. Lalos, E.~K. Markakis, and I.~Politis, ``A federated deep unrolling method for lidar super-resolution: Benefits in slam,'' \emph{IEEE Transactions on Intelligent Vehicles}, vol.~9, no.~1, pp. 199--215, 2024.

\bibitem{10167942}
A.~Gkillas, G.~Arvanitis, A.~S. Lalos, and K.~Moustakas, ``Federated learning for lidar super resolution on automotive scenes,'' in \emph{2023 24th International Conference on Digital Signal Processing (DSP)}, 2023, pp. 1--5.

\bibitem{10167897}
C.~Anagnostopoulos, A.~Gkillas, N.~Piperigkos, and A.~S. Lalos, ``Federated deep feature extraction-based slam for autonomous vehicles,'' in \emph{2023 24th International Conference on Digital Signal Processing (DSP)}, 2023, pp. 1--5.

\bibitem{9981098}
L.~Fantauzzo, E.~Fanì, D.~Caldarola, A.~Tavera, F.~Cermelli, M.~Ciccone, and B.~Caputo, ``Feddrive: Generalizing federated learning to semantic segmentation in autonomous driving,'' in \emph{2022 IEEE/RSJ International Conference on Intelligent Robots and Systems (IROS)}, 2022, pp. 11\,504--11\,511.

\bibitem{shi_beyond_2022}
\BIBentryALTinterwordspacing
Y.~Shi and H.~Li, ``Beyond {Cross}-view {Image} {Retrieval}: {Highly} {Accurate} {Vehicle} {Localization} {Using} {Satellite} {Image},'' Sep. 2022, arXiv:2204.04752 [cs]. [Online]. Available: \url{http://arxiv.org/abs/2204.04752}
\BIBentrySTDinterwordspacing

\bibitem{vgg}
\BIBentryALTinterwordspacing
K.~Simonyan and A.~Zisserman, ``Very deep convolutional networks for large-scale image recognition,'' 2014. [Online]. Available: \url{https://arxiv.org/abs/1409.1556}
\BIBentrySTDinterwordspacing

\bibitem{LM}
\BIBentryALTinterwordspacing
K.~LEVENBERG, ``A method for the solution of certain non-linear problems in least squares,'' \emph{Quarterly of Applied Mathematics}, vol.~2, no.~2, pp. 164--168, 1944. [Online]. Available: \url{http://www.jstor.org/stable/43633451}
\BIBentrySTDinterwordspacing

\bibitem{mcmahan2023communicationefficientlearningdeepnetworks}
\BIBentryALTinterwordspacing
H.~B. McMahan, E.~Moore, D.~Ramage, S.~Hampson, and B.~A. y~Arcas, ``Communication-efficient learning of deep networks from decentralized data,'' 2023. [Online]. Available: \url{https://arxiv.org/abs/1602.05629}
\BIBentrySTDinterwordspacing

\bibitem{Geiger2013IJRR}
A.~Geiger, P.~Lenz, C.~Stiller, and R.~Urtasun, ``Vision meets robotics: The kitti dataset,'' \emph{International Journal of Robotics Research (IJRR)}, 2013.

\end{thebibliography}

\end{document}